\documentclass[letterpaper, 10 pt, journal, twoside]{IEEEtran}  

\IEEEoverridecommandlockouts                              

\makeatletter

\newcommand{\Rmnum}[1]{\expandafter\@slowromancap\romannumeral #1@}
\makeatother

\usepackage{amsmath} 
\usepackage{amssymb}
\usepackage{amsthm}
\usepackage{verbatim}

\usepackage[linesnumbered,ruled]{algorithm2e}
\usepackage{latexsym}

\usepackage{bm}

\usepackage{graphicx}
\usepackage{subfig}
\usepackage{xcolor}

\newtheorem{assumption}{Assumption}
\newtheorem{definition}{Definition}
\newtheorem{problem}{Problem}
\newtheorem{theorem}{Theorem}
\begin{document}

\title{Chance-Constrained Iterative Linear-Quadratic Stochastic Games}

\author{Hai Zhong$^{1}$, ~\IEEEmembership{Student Member,~IEEE}, Yutaka Shimizu$^{2}$, and Jianyu Chen*$^{3}$, ~\IEEEmembership{Member,~IEEE}%
\thanks{Manuscript received: July, 8, 2022; Revised October, 5, 2022; Accepted November, 22, 2022.}
\thanks{This paper was recommended for publication by Editor M. Ani Hsieh upon evaluation of the Associate Editor and Reviewers' comments.} 
\thanks{$^{1}$ Hai Zhong  is with the Institute for Interdisciplinary Sciences, Tsinghua University, Beijing, China.
        {\tt\footnotesize 
        zhongh22@mails.tsinghua.edu.cn}}%
\thanks{$^{2} $ Yutaka Shimizu is with  Tier IV, Inc., Jacom Building, 1-12-10 Kitashinagawa, Shinagawa-ku, Tokyo, 140-0001, Japan.
        {\tt\footnotesize
        purewater0901@gmail.com}}%
\thanks{$^{3} $ Jianyu Chen is with the Institute for Interdisciplinary Sciences, Tsinghua University, Beijing, China, and the Shanghai Qizhi Institute, Shanghai, China (*corresponding author).
        {\tt\footnotesize
        jianyuchen@tsinghua.edu.cn}}%
\thanks{Digital Object Identifier (DOI): see top of this page.}
}

\markboth{IEEE Robotics and Automation Letters. Preprint Version. Accepted November, 2022}
{Zhong \MakeLowercase{\textit{et al.}}: Chance-Constrained Iterative Linear-Quadratic Stochastic Games}

\IEEEaftertitletext{\vspace{-1\baselineskip}}
\maketitle

\begin{abstract}
Dynamic game arises as a powerful paradigm for multi-robot planning, for which safety constraint satisfaction is crucial. Constrained stochastic games are of particular interest, as real-world robots need to operate and satisfy constraints under uncertainty. Existing methods for solving stochastic games handle chance constraints using exponential penalties with hand-tuned weights. However, finding a suitable penalty weight is nontrivial and requires trial and error. In this letter, we propose the chance-constrained iterative linear-quadratic stochastic games (CCILQGames) algorithm. CCILQGames solves chance-constrained stochastic games using the augmented Lagrangian method. We evaluate our algorithm in three autonomous driving scenarios, including merge, intersection, and roundabout. Experimental results and Monte Carlo tests show that CCILQGames can generate safe and interactive strategies in stochastic environments.
\end{abstract}


\section{INTRODUCTION}

\IEEEPARstart{W}{ith} the recent advances in multi-robot systems, dynamic game has emerged as a new paradigm for multi-robot planning \cite{bacsar1998dynamic,ILQGames}. Dynamic game naturally captures the interactive nature of multi-agent planning, as the ego agent's strategy accounts for the fact that other agents' plans could change based on the ego agent's action. A few works also consider safety constraints in the dynamic game settings \cite{ALgames,di2020local}, which is another key aspect of multi-robot planning (e.g., avoiding collisions with other robots or obstacles).

\begin{figure}[htbp]
    \centering
    \subfloat[]{\includegraphics[width=0.25\textwidth]{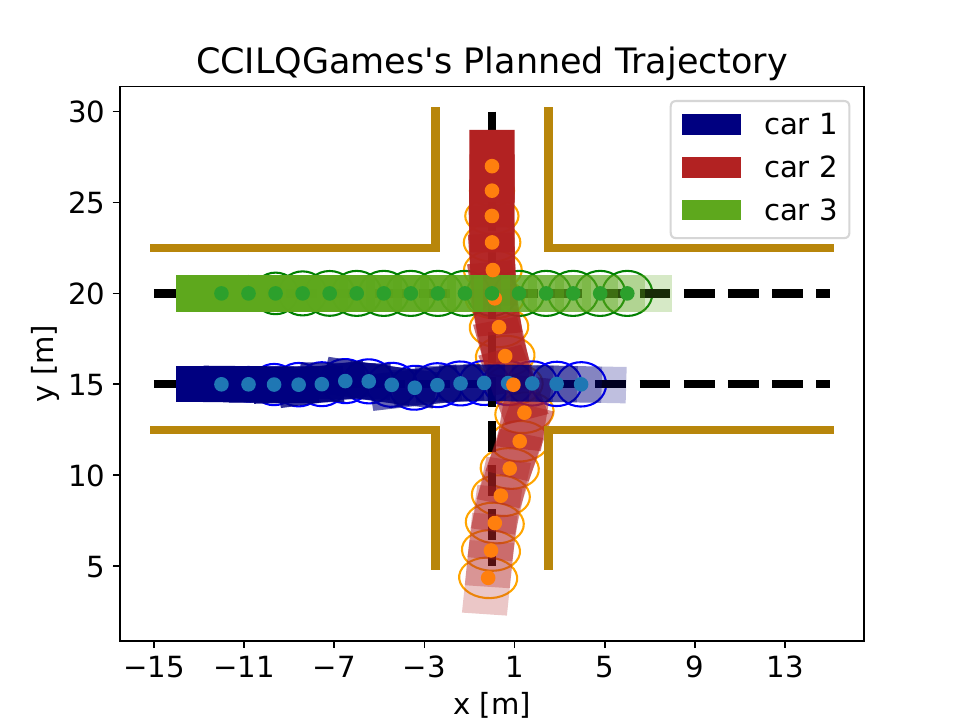}} 
    \subfloat[]{\includegraphics[width=0.25\textwidth]{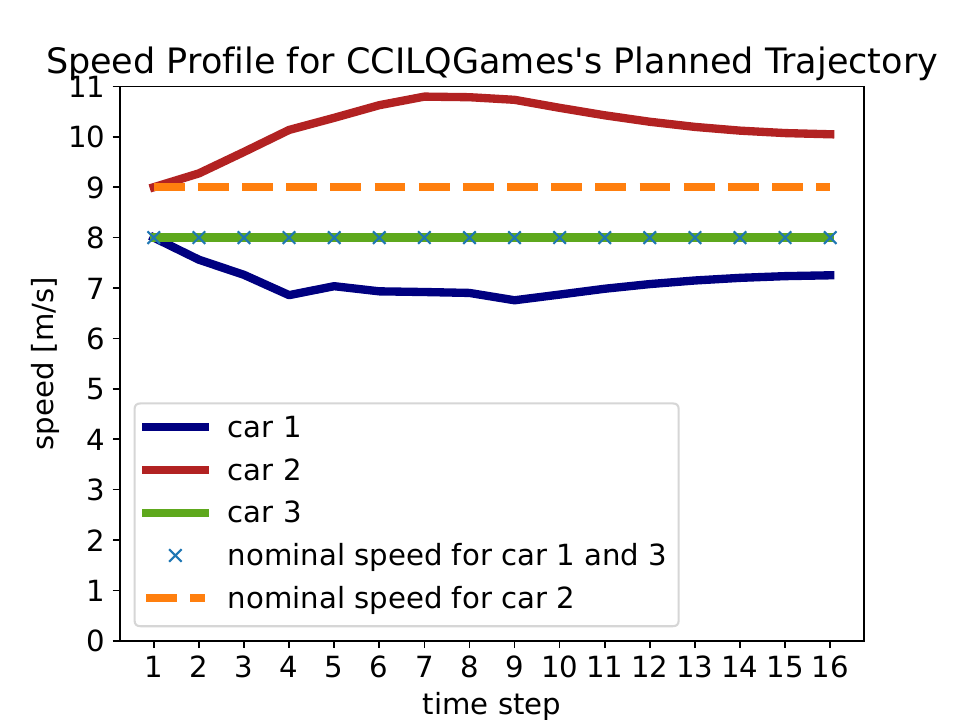}} 
    \quad
    \subfloat[]{\includegraphics[width=0.25\textwidth]{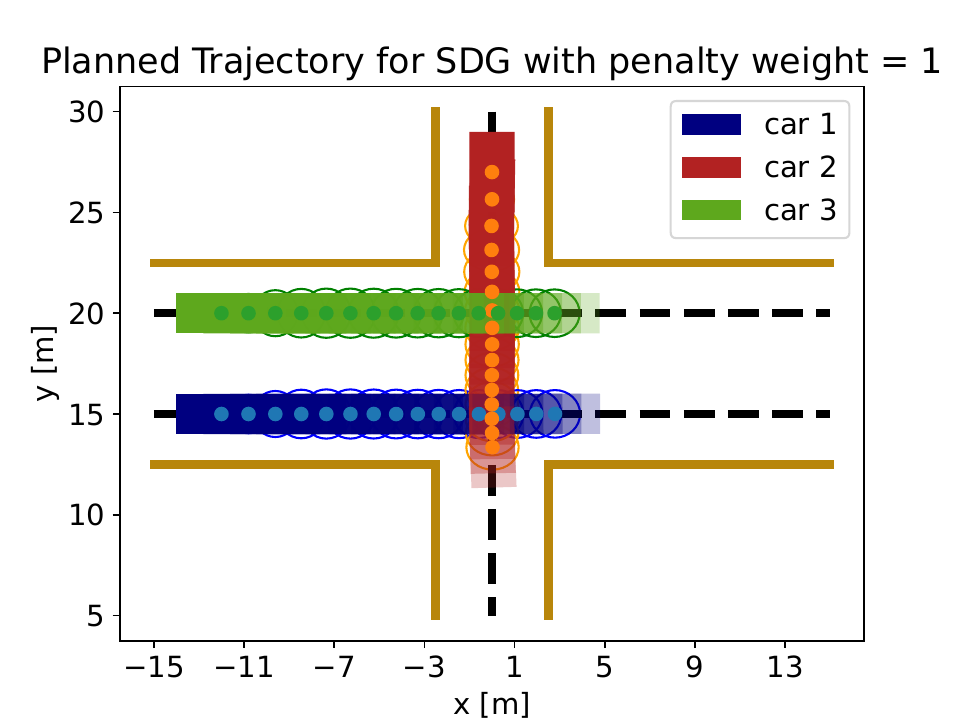}}
    \subfloat[]{\includegraphics[width=0.25\textwidth]{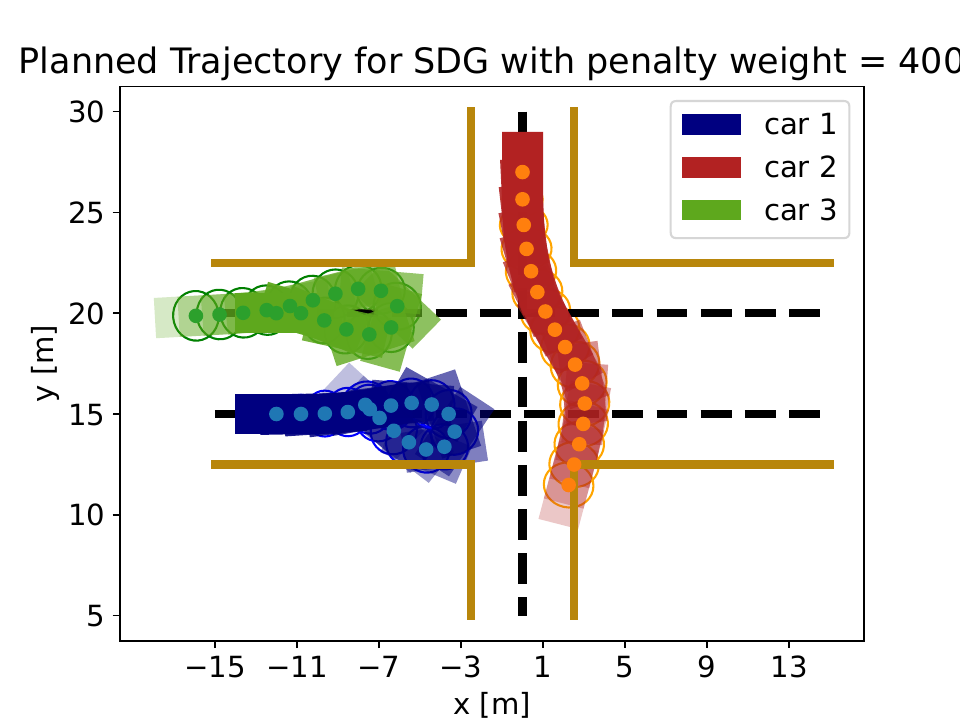}}
    \caption{SDG with a large penalty weight could lead to conservative strategies (as shown in (d)), while a small penalty weight could cause unsafe trajectories (as shown in (c)). CCILQGames could ensure performance while satisfying constraints without manually tuning penalty weights (as shown in (a)).}
    \label{fig:Motivating example}
\end{figure}

Most of the existing work on dynamic games focuses on deterministic settings. However, robots operating in the real world must reason about uncertainties, resulting in stochastic games. For example, robots need to reason about their unmodeled or disturbed system dynamics, noisy sensor perceptions, and other agents' motion and intention uncertainties. Also, the complexity of satisfying safety constraints in stochastic games is further evolved, as the agent needs to guarantee safety under uncertainty. Schwarting et al. \cite{StochasticDynamicGames} recently proposed the stochastic dynamic game (SDG) algorithm to solve stochastic games, considering process and observation uncertainties. SDG simply adds exponential penalties for chance constraint violations. Yet, this approach is far from being satisfactory: choosing the penalty weight is nontrivial. Inappropriate penalty weights will result in undesired behaviors: a small penalty leads to dangerous collisions (Figure \ref{fig:Motivating example}(c)), while a large penalty leads to conservative suboptimal behaviors (Figure \ref{fig:Motivating example}(d)). Tuning penalty terms requires trial and error and is generally failure-prone, which prohibits the potential for real-time applications. To the best of our knowledge, no existing work systematically handles safety constraints for stochastic dynamic games (with both process and observation uncertainties).

This letter proposes the chance-constrained iterative linear-quadratic stochastic games (CCILQGames) algorithm to overcome the hurdle above. CCILQGames can incorporate chance constraints in stochastic dynamic games, which allows specification of the probability threshold for successfully satisfying constraints. We use the augmented Lagrangian method to construct an outer and inner loop framework, with the benefits of automatically tuning the penalty weights. The chance constraint violation is calculated to update penalty and Lagrange multiplier terms in the outer loop. We solve the corresponding unconstrained stochastic game in the inner loop by extending the iterative linear-quadratic games (ILQGames) algorithm \cite{ILQGames,ILQGamesFeedbackLinearization} to the stochastic game setting. We evaluate the proposed method in lane merging, three-player intersection, and roundabout scenarios. Experimental results and Monte Carlo tests show that our approach generates interactive and safe strategies under uncertainty.

\section{RELATED WORK}
\subsection{Game-Theoretical Planning}
A family of algorithms for solving differential games hinges on transforming a differential game into one or a series of single-agent optimal control problems. Kavuncu et al. \cite{PotentialILQR} pursue this idea by utilizing potential differential games, for which Nash equilibria could be obtained by solving an associated optimal control problem. Iterative best response scheme is another family of algorithms that takes advantage of tools for solving optimal control problems \cite{BestMultiRacing,IterativeBestCars}. This approach solves Nash equilibria by solving single-agent optimal control problems in a round-robin fashion. The usefulness of this type of algorithm is demonstrated in racing games \cite{BestMultiRacing,IterativeBestCars}. A pitfall for this approach is that there are no formal guarantees for convergence to Nash equilibria. 

Another thrust of game-theoretic solvers focuses on solving for open-loop Nash equilibrium. The key to this type of approach is to formulate the necessary conditions of open-loop Nash equilibria by concatenating the Karush–Kuhn–Tucker conditions for each player. Di et al. \cite{DDPGameDI} solved the corresponding nonlinear program based on Newton's method. In \cite{ALgames}, augmented Lagrangian is combined with Newton's method to handle state and control constraints. In \cite{di2020local}, the constrained nonlinear program is addressed by projected gradient or Douglas-Rachford splitting method. As this type of approach solves for open-loop Nash equilibrium (i.e., controls are functions of time instead of states), special treatments such as model predictive control are required to help to capture the reactive game-theoretic nature \cite{ALgames}. Another limitation of the aforementioned approaches is that they only consider deterministic games and can not handle uncertainties in dynamics and observations.

An increasingly popular branch of game-theoretic solvers aims to compute local feedback Nash equilibrium \cite{ILQGames,ILQGamesFeedbackLinearization,DDPGameDI,StochasticDynamicGames,laine2021computation}. The iterative linear-quadratic game method \cite{ILQGames} exploits the analytical solutions of linear-quadratic games and solves the game as a sequence of approximated linear-quadratic games. Schwarting et al. \cite{StochasticDynamicGames} propose a belief space variant of differential dynamic programming algorithm for stochastic games, which could handle dynamics and observation uncertainties. However, their work handles chance constraints using exponential penalties, which requires adjusting penalty weights by trial and error. Our work builds upon the iterative linear-quadratic game method and extends to stochastic games with dynamics and observation uncertainty. In contrast to \cite{StochasticDynamicGames}, our work directly handles chance constraints using the augmented Lagrangian method. Our approach does not require manually adjusting penalty weights and achieves a good balance of safety and performance.

\subsection{Chance-Constrained Planning}
Chance constraint formulation provides a unified framework for quantifying probabilistic constraint violations, which allows the specification of desired probabilities for constraint violations. However, calculating chance constraint violations is intractable in general, which makes chance-constrained planning challenging. 
A branch of works exploits the structure of Gaussian belief distributions \cite{StochasticPredictiveControl,ChanceConstrainedILQG}, whether it be a true belief distribution or an approximation.  We follow this line of work, exploiting the structure of linear-Gaussian systems via linearization.  We further incorporate this technique into an augmented Lagrangian framework. A different line of research exploits moment-based approaches to handle non-Gaussian uncertainties \cite{MomentBasedNonGaussianRisk}. \cite{MomentBasedNonGaussianRisk} proposed a moment-based method that can handle general chance-constrained motion planning problems. The key to handling general chance-constrained planning problems is to transform chance constraints into a set of deterministic constraints on the moments of the probability distributions of the states, along with an exact moment-based uncertainty propagation method to relate the moments of the uncertain states to control inputs.
\section{PROBLEM FORMULATION}

We consider the following nonlinear stochastic $N$-player dynamics:
\begin{equation} \label{eq:nonlinear_dynamics_model}
\begin{aligned}
x_{k+1}  = f(x_k,u_k^{1:N},w_k), \quad w_k  \sim \mathcal{N}(0,\Sigma_{w_k}),\\
\end{aligned}
\end{equation}
where $x_k$ and $x_{k+1} \in \mathbb{R}^{n}$ are the states at time step $k$ and $k+1$. $u_k^{i} \in \mathbb{R}^{m_i}, i \in \{1,...,N\}$ is the control input of player $i$ at time step $k$, and $u_k^{1:N} = \{u_k^{1},...,u_k^{N}\}$ is the collection of all players' controls at time step $k$. $w_k \in \mathbb{R}^{d}$ represents the Gaussian process noise at time step $k$. $\Sigma_{w_k}$ is the corresponding covariance for $w_k$. Although the dynamics model considers Gaussian noises, it is still general as the Gaussian noises could go through arbitrary nonlinear transformations through $f$. 

Following \cite{StochasticDynamicGames}, we assume that each agent shares the same observation model and state measurement. At the core of CCILQGames is solving a series of linear-quadratic stochastic games. Linear-quadratic stochastic games with certain structures are computationally tractable \cite{StochasticGameConstrainedEstimators,1980StochasticGame}, thanks to the separation principle. In this work, assumption 1 is imposed to make the linear-quadratic stochastic games tractable. We state this assumption formally below:
\begin{assumption}
All agents share the same joint measurement model and receive the same measurement of the state (i.e., $x_k$) at each stage, generated by a common observation-noise realization (which cannot be directly observed). At stage $k$, past control inputs are common knowledge for all agents.\footnote{In this updated version, Assumption 1 and the strategy space definition have been revised to make the information structure explicit. The original version considered the same strategy class, but the dependence on past control inputs was not explicitly stated in the strategy space definition, despite being implicitly used in the Kalman filter recursion.}
\end{assumption}

With assumption 1, we consider the following measurement model:
\begin{equation} \label{eq:nonlinear_observation_model}
\begin{aligned}
y_{k+1}  = h(x_{k+1},v_{k+1}), \quad v_{k+1}  \sim \mathcal{N}(0,\Sigma_{v_{k+1}}), \\
\end{aligned}
\end{equation}
where $y_{k+1} \in \mathbb{R}^{z}$ and  $v_{k+1} \in \mathbb{R}^{s}$  are the sensor measurement and the Gaussian measurement noise at time step $k+1$. $\Sigma_{v_{k+1}}$ is the covariance of the measurement noise. 

System dynamics $(1)$ and measurement model $(2)$ jointly define a partially observable Markov decision process, in which agents maintain a probability distribution of the state $x_k$. We use belief $b_k$ to represent the probability distribution of state $x_k$ conditioned on control inputs and measurements:
\begin{equation}
\begin{aligned}
b_k  = Pr(x_k|y_0,...,y_k,u_0^{1:N},...,u_k^{1:N}). \\
\end{aligned}
\end{equation}

Furthermore, we consider chance constraints on state:
\begin{equation}
\begin{aligned}
Pr(g_{k}(x_k)\leq 0) \geq p, \\
\end{aligned}
\end{equation}
where $g_{k}(x_k)\leq 0$ is the nonlinear constraint on state $x_k$ at time step $k$. $p$ is the chance constraint threshold. We assume $g_{k}(x_k)$ is a scalar.

Each agent has a deterministic time-varying feedback control policy $\gamma_i(k, y_0,...,y_k, u_0^{1:N},...,u_{k-1}^{1:N}) \in \Gamma_i$, which maps the current time step, observations, and collective past control inputs to a control input (i.e., $u_k^{i} = \gamma_i(k, y_0,...,y_k, u_0^{1:N},...,u_{k-1}^{1:N})$). Note that at each stage, each agent has access to the control inputs applied at previous stages by assumption. $\Gamma_i$ is the strategy space for player $i$. Each agent has a cost function $J_i$ defined as the expectation of states and measurements:
\begin{equation} \label{eq:agent_cost_function}
\begin{aligned}
J_i(\gamma_1,...,\gamma_N) & = \\ \mathop{\mathbb{E}}_{X_0,...,X_L,Y_0,...,Y_L}& \left[c_L^{i}(x_L) +\sum_{k=0}^{L-1} c_k^{i}(x_k,u_k^{1:N})\right],\\
\end{aligned}
\end{equation}
where $L$ is the planning horizon, and $X_0,...,X_L,Y_0,...,Y_L$ are random variables for states and measurements. Now we are ready to define the generalized Nash equilibrium, which is the  Nash equilibrium under chance constraints:
\begin{definition}
A set of control policies $\{\gamma_i^{*}\}, i \in \{1,...,N\}$, is a generalized Nash equilibrium if the following inequality holds for all players while satisfying chance constraints (4):
\begin{equation}
\begin{aligned}
J_i(\gamma_1^{*},...,\gamma_i^{*},...,\gamma_N^{*}) & \leq J_i(\gamma_1^{*},...,\gamma_i,...,\gamma_N^{*}), \\
&\forall \gamma_i \in \Gamma_i, i \in \{1,...,N\}.\\
\end{aligned}
\end{equation}
\end{definition}

As we describe in detail in section \Rmnum{4}.C, our work builds upon ILQGames. Since the computation of a Nash equilibrium is generally intractable, ILQGames attempts to find local Nash equilibria (i.e., the inequality in equation (6) holds within a neighborhood around each player's local Nash equilibrium strategy) \cite[Definition 1]{ChracaterizationOfLocalNash}. ILQGames solves a series of linear-quadratic game approximations to the original game. More precisely, ILQGames obtains the feedback Nash equilibrium \cite[Definition 6.2]{bacsar1998dynamic} solution to the approximate linear-quadratic game. The feedback Nash equilibrium solution is strongly time-consistent \cite[Chapter 6, Theorem 6.6]{bacsar1998dynamic}, which is appealing since the feedback Nash equilibrium strategy remains optimal at a future time step even if there are deviations from the feedback strategy in the past time steps. Given that ILQGames is at the core of our algorithm, CCILQGames also aims to find local Nash equilibrium in practice.

With the above components, we can finally derive the target problem for this work (i.e., finding the generalized Nash equilibrium), which is summarized as follows:
\begin{problem}
Solve for a generalized Nash equilibrium with the following cost functions, dynamics model, measurement model, and chance constraints:
\begin{equation*}
\begin{aligned}
J_i(\gamma_1,...,\gamma_N)  &=\mathop{\mathbb{E}}[c_{L}^{i}(x_L) +\sum_{k=0}^{L-1} c_k^{i}(x_k,u_k^{1:N})],\\
x_{k+1}  &= f(x_k,u_k^{1:N},w_k), \quad w_k  \sim \mathcal{N}(0,\Sigma_{w_k}),\\
y_{k+1}  &= h(x_{k+1},v_{k+1}), \quad v_{k+1}  \sim \mathcal{N}(0,\Sigma_{v_{k+1}}), \\
Pr(g_{k}^{i,m}&(x_k)\leq 0) \geq p_{i,m}, \quad m \in \{1,..,M_k^{i}\}, \\
i &\in \{1,...,N\},\\
\end{aligned}    
\end{equation*}
\end{problem}
\noindent where  $m$, $p_{i,m}$, and $M_k^{i}$ are the index for the $m^{th}$ chance constraint, chance constraint threshold and the total number of constraints at time step $k$ for agent $i$.

\section{CHANCE-CONSTRAINED ITERATIVE LINEAR-QUADRATIC GAME}
\subsection{System Linearization and Belief Dynamics}
Given the current belief $b_k$ and next time step's measurement $y_{k+1}$, the belief dynamics could be described using the Bayesian filter \cite{ProbabilisticRobotics}. Following \cite{LinearizationILQG, ChanceConstrainedILQG}, we linearize the dynamical system around a nominal trajectory and then apply the Kalman filter to approximately track the belief dynamics. Kalman filter approximates the true belief as a Gaussian distribution, such that $b_k = (\hat{x}_k,\Sigma_{x_k})$, where $\hat{x}_k$ is the mean and $\Sigma_{x_k}$ is the covariance. 

Given a nominal trajectory  $\bar{b} =\{\bar{b}_k = (\bar{x}_k,\bar{\Sigma}_{x_k}), k \in \{0,...,L\}\},\bar{u}_{0}^{1:N},...,\bar{u}_{L-1}^{1:N}$, satisfying $\bar{x}_{k+1} = f(\bar{x}_k,\bar{u}_k^{1:N},0)$, we linearize the dynamics~\eqref{eq:nonlinear_dynamics_model} and measurement models \eqref{eq:nonlinear_observation_model}:
\begin{equation} \label{eq:linearized_dynamics_and_observation_model}
\begin{aligned}
x_{k+1}  &= \bar{x}_{k+1}+A_k(x_k-\bar{x}_k)+\sum_{j=1}^{N} B_k^j(u^{j}_{k}-\bar{u}^{j}_{k})+W_{k}w_{k},\\
y_{k+1}  &= h(\bar{x}_{k+1},0) + H_{k+1}(x_{k+1}-\bar{x}_{k+1}) + V_{k+1}v_{k+1},\\
\end{aligned}    
\end{equation}
where $A_k, B_k^j, W_{k}, H_{k+1}, V_{k+1}$ are:
\begin{equation}
\begin{aligned}
A_{k}  &= \frac{\partial f(\bar{x}_k,\bar{u}_k^{1:N},0)}{\partial x_k},\quad & B_{k}^{j}  &= \frac{\partial f(\bar{x}_k,\bar{u}_k^{1:N},0)}{\partial {u}_k^{j} },\\
W_{k}  &= \frac{\partial f(\bar{x}_k,\bar{u}_k^{1:N},0)}{\partial w_k},\quad & H_{k+1}  &= \frac{\partial h(\bar{x}_{k+1},0)}{\partial x_{k+1}},\\
V_{k+1} &= \frac{\partial h(\bar{x}_{k+1},0)}{\partial v_{k+1}}.
\end{aligned}    
\end{equation}

Now we can use Kalman filter to propagate next time step's belief $b_{k+1} = (\hat{x}_{k+1},\Sigma_{x_{k+1}})$ with the linearized dynamics, observation model, and current belief $b_k = (\hat{x}_k,\Sigma_{x_k})$:
\begin{equation}
\begin{aligned}
\hat{x}_{k+1}^{p} &= \bar{x}_{k+1}+A_k(\hat{x}_k-\bar{x}_k)+\sum_{j=1}^{N} B_k^j(u^{j}_{k}-\bar{u}^{j}_{k}),\\
\Sigma_{x_{k+1}}^{p} &= A_k \Sigma_{x_k} A_k^{T} + W_k \Sigma_{w_k} W_k^{T},\\
\hat{x}_{k+1} &= \bar{x}_{k+1}+A_k(\hat{x}_k-\bar{x}_k)+\sum_{j=1}^{N} B_k^j(u^{j}_{k}-\bar{u}^{j}_{k})\\
+&K_{k+1}(y_{k+1}-(h(\bar{x}_{k+1},0) + H_{k+1}(\hat{x}_{k+1}^{p}-\bar{x}_{k+1}))), \\
\Sigma_{x_{k+1}} &= (I-K_{k+1}H_{k+1})\Sigma_{x_{k+1}}^{p},\\
K_{k+1} = &\Sigma_{x_{k+1}}^{p}H_{k+1}^{T}(H_{k+1}\Sigma_{x_{k+1}}^{p}H_{k+1}^{T}+V_{k+1}\Sigma_{v_{k+1}}V_{k+1}^{T})^{-1},\\
\end{aligned}    
\end{equation}
where $\hat{x}_{k+1}^{p}$ is the state's prior estimation, $\Sigma_{x_{k+1}}^{p}$ is the covariance's prior estimation, and $K_{k+1}$ is the Kalman gain. The innovation term $y_{k+1}-(h(\bar{x}_{k+1},0) + H_{k+1}(\hat{x}_k-\bar{x}_k))$ is random, which renders the mean propagation stochastic. Although the mean propagation is stochastic, the covariance propagation is deterministic, dependent neither on measurements nor controls.


\subsection{Outer Loop: Augmented Lagrangian Scheme}
\subsubsection{Augmented Lagrangian Formulation}
The augmented Lagrangian associated with player $i$ is formulated by augmenting the cost function~\eqref{eq:agent_cost_function} with Lagrange multiplier terms and quadratic penalty terms:
\begin{equation}
\begin{aligned}
&l_{i}(X_0,...,X_L,Y_0,...,Y_L,u^{1:N}_{0},...,u^{1:N}_{L-1})\\
&=\mathop{\mathbb{E}}_{X_0,...,X_L,Y_0,...,Y_L}[c_{L}^{i}(x_L) +\sum_{k=0}^{L-1} c_k^{i}(x_k,u_k^{1:N})]\\
&+\sum_{k=1}^{L}\sum_{m=1}^{M_k} \lambda_{i,m}[p_{i,m}-Pr(g_{k}^{i,m}(x_k) \leq 0)]\\
&+\sum_{k=1}^{L}\sum_{m=1}^{M_k} \frac{I_{i,m}}{2}[p_{i,m}-Pr(g_{k}^{i,m}(x_k) \leq 0)]^2,\\
\end{aligned}    
\end{equation}
where $\lambda_{i,m} \in \mathbb{R}$ is the Lagrange multiplier. $I_{i,m}$ is defined as:
\begin{equation}
\begin{aligned}
I_{i,m} = \begin{cases}
          0 ,  \quad &p_{i,m}-Pr(g_{k}^{i,m}(x_k) \leq 0) < 0 \wedge \lambda_{i,m} =0, \\
          \mu_{i,m}, \quad & otherwise. \\
     \end{cases}
\end{aligned}    
\end{equation}
where $\mu_{i,m} \in \mathbb{R}$ is the quadratic penalty term. 
\subsubsection{Augmented Lagrangian Update}
Given a trajectory, the Lagrange multipliers and penalty terms are updated as follows \cite{Nocedal2006Numerical}:
\begin{equation}
\begin{aligned}
\lambda_{i,m} &= \max(0,\lambda_{i,m} + \mu_{i,m} (p_{i,m}-Pr(g_{k}^{i,m}(x_k) \leq 0)   ), \\
\mu_{i,m} &= \phi \mu_{i,m}, \\
\end{aligned}    
\end{equation}
where $\phi > 1$ is the increasing schedule.
\subsubsection{Evaluating Chance Constraints Violations}
Evaluating chance constraint violations is necessary to update the Lagrange multipliers. However, directly calculating chance constraint violations could be intractable in general. Hence, we follow the approach of \cite{StochasticPredictiveControl,ChanceConstrainedILQG} to linearize chance constraints. 

Given a trajectory with belief $\bar{b}_k = (\bar{x}_k,\bar{\Sigma}_{x_k})$ and control $\bar{u}_k^{1:N}$, we get a linearization of $g_{k}^{i,m}(x_k)$ around $\bar{x}_k$:
\begin{equation}
\begin{aligned}
G_k^{i,m} x_k + q_k^{i,m},\\
\end{aligned}    
\end{equation}
where 
\begin{equation}
\begin{aligned}
G_k^{i,m} = \frac{\partial g_{k}^{i,m}(\bar{x}_k)}{\partial x_k}, \quad q_k^{i,m} = g_{k}^{i,m}(\bar{x}_k) - G_k^{i,m} \bar{x}_k.\\
\end{aligned}    
\end{equation}

Now we can consider the following linearized chance constraints:
\begin{equation} \label{eq:linearized_chance_constraint}
\begin{aligned}
Pr(  G_k^{i,m} x_k + q_k^{i,m} \leq 0         ) \geq p_{i,m}. \\
\end{aligned}
\end{equation}

To step further, we can decompose the Gaussian random variable $x_k$ into its mean $\bar{x}_k$ and $\bar{e}_k = {x}_k - \bar{x}_k  \sim \mathcal{N}(0,\bar{\Sigma}_{x_k})$:
\begin{equation}
\begin{aligned}
Pr( G_k^{i,m} (\bar{x}_k+\bar{e}_k) + q_k^{i,m} \leq 0         ) \geq p_{i,m}. \\
\end{aligned}
\end{equation}

By rearranging terms, chance constraint~\eqref{eq:linearized_chance_constraint} can be transformed to a deterministic linear constraint on mean $\bar{x}_k$ \cite{StochasticPredictiveControl}:
\begin{equation} \label{eq:deterministic_chance_constraint}
\begin{aligned}
G_k^{i,m} \bar{x}_k \leq  - q_k^{i,m} - \bar{\rho}_k^{i,m}.\\
\end{aligned}
\end{equation}

$\bar{\rho}_k^{i,m}$ could be calculated using the quantile function for univariate Gaussian, as $G_k^{i,m}\bar{e}_k \sim \mathcal{N}(0,G_k^{i,m}\bar{\Sigma}_{x_k}(G_k^{i,m})^{T})$:
\begin{equation}
\begin{aligned}
Pr&( G_k^{i,m} e_k \leq   \bar{\rho}_k^{i,m}       ) = p_{i,m},\\
\bar{\rho}_k^{i,m} &= \sqrt{2G_k^{i,m}\bar{\Sigma}_{x_k}(G_k^{i,m})^{T}} erf^{-1} (2p_{i,m}-1),\\
\end{aligned}
\end{equation}
where $erf^{-1}$ is the inverse error function. Since mean $\bar{x}_k$ and covariance $\bar{\Sigma}_{x_k}$ are known for the nominal trajectory, the violation of constraint~\eqref{eq:deterministic_chance_constraint} can be directly calculated.


\subsection{Inner Loop: Iterative Linear Quadratic Game}
The inner loop iteratively solves a linear-quadratic stochastic game. We present how to construct the approximate linear-quadratic game from the original game and solve it.
\subsubsection{Approximate Linear-Quadratic Game}
The inner loop of the proposed CCILQGames performs an iterative linear-quadratic stochastic game step, which involves iteratively solving a linear-quadratic stochastic game. We now present how to obtain an approximated linear-quadratic stochastic game with a given nominal trajectory $\bar{b} =\{b_k = (\bar{x}_k,\bar{\Sigma}_{x_k}), k \in \{0,...,L\}\}$, control strategies $\{\bar{\gamma_i}\}, i \in \{1,...,N\}$ and controls $\bar{u}_{0}^{1:N},...,\bar{u}_{L-1}^{1:N}$.

Toward this end, we first obtain the linearized dynamics and observation model~\eqref{eq:linearized_dynamics_and_observation_model} around the nominal trajectory, as described in section \Rmnum{4}.A.

Moreover, a quadratic approximation to the running costs $c_k^{i}(x_k,u_k^{1:N})$ is necessary. Yet, before we are able to proceed, we need to put the Lagrange multiplier and penalty terms associated with chance constraints inside the expectation operator.

We proceed in the same manner as in section \Rmnum{4}.B.(3). By linearizing around the nominal trajectory, we obtain a linear approximation to chance constraints in the form of a deterministic linear constraint on the mean $\hat{x}_k$:
\begin{equation}
\begin{aligned}
G_k^{i,m} &\hat{x}_k \leq  - q_k^{i,m} - \rho_k^{i,m},\\
\rho_k^{i,m} &= \sqrt{2G_k^{i,m}{\Sigma}_{x_k}(G_k^{i,m})^{T}} erf^{-1} (2p_{i,m}-1),\\
\end{aligned}
\end{equation}
where $q_k^{i,m}$ is defined in (14). Since the covariance propagation is deterministic as we mentioned in Section IV.A, we can precompute the covariance before solving the game.

The linear constraint (19) now replaces the original chance constraint. The Lagrange multiplier term can be directly put inside the expectation operator:
\begin{equation}
\begin{aligned}
&\lambda_{i,m}(G_k^{i,m} \hat{x}_k  + q_k^{i,m} + \rho_k^{i,m}) \\ &=\mathop{\mathbb{E}}_{X_k}[\lambda_{i,m}(G_k^{i,m} {x}_k  + q_k^{i,m} + \rho_k^{i,m})].\\
\end{aligned}    
\end{equation}

For any probability distribution of $X_k$, $I_{i,m}$ is determined since the mean $\hat{x}_k$ is determined. Now, the penalty term can be transformed as follows:
\begin{equation}
\begin{aligned}
&\frac{I_{i,m}}{2}(G_k^{i,m} \hat{x}_k + q_k^{i,m} + \rho_k^{i,m})^{2}\\
&=\frac{I_{i,m}}{2} [\hat{x}_k^{T}{G_k^{i,m}}^{T}G_k^{i,m} \hat{x}_k +2(q_k^{i,m} + \rho_k^{i,m})^{T}G_k^{i,m} \hat{x}_k\\
&+(q_k^{i,m} + \rho_k^{i,m})^{T}(q_k^{i,m} + \rho_k^{i,m}) ]\\
&=\frac{I_{i,m}}{2}\mathop{\mathbb{E}}_{X_k}[{x}_k^{T}{G_k^{i,m}}^{T}G_k^{i,m} {x}_k+2(q_k^{i,m} + \rho_k^{i,m})^{T}G_k^{i,m} {x}_k\\
&+(q_k^{i,m} + \rho_k^{i,m})^{T}(q_k^{i,m} + \rho_k^{i,m}) ]\\& -\frac{I_{i,m}}{2}trace({G_k^{i,m}}^{T}G_k^{i,m}\Sigma_{x_k}),\\
\end{aligned}    
\end{equation}
Since ${G_k^{i,m}}$ and $\Sigma_{x_k}$ are constants after linearization around the nominal trajectory, $trace({G_k^{i,m}}^{T}G_k^{i,m}\Sigma_{x_k})$ is also constant. But the penalty $I_{i,m}$ indeed depends on the control, as the control affects the probability distribution of $X_k$. So when constructing the linear-quadratic game, the trace term $trace({G_k^{i,m}}^{T}G_k^{i,m}\Sigma_{x_k})$ should be taken into account. However, calculating the derivative and hessian of it requires calculating second and third order derivatives of the dynamics, which could add heavy computational costs. So we omit the trace term in each agent's augmented Lagrangian as an approximation in this work.

Combining $(20)$ and $(21)$, the Lagrange  multiplier and penalty terms are transformed into the expectation operator. We denote the modified running cost as $\Tilde{c_k}^{i}(x_k,u_k^{1:N})$. Finally, we are equipped to obtain a quadratic approximation for the running costs:
\begin{equation}
\begin{aligned}
\Tilde{c_k}^{i}(x_k,u_k^{1:N}) &\approx \Tilde{c_k}^{i}(\bar{x}_k,\bar{u}_k^{1:N})+\frac{1}{2} \delta x_k^{T}(Q_k^{i}\delta x_k+2l_k^{i}) \\
&+\frac{1}{2} \sum_{j=1}^{N} \delta {u_k^{j}}^{T}(R_k^{ij}\delta u_k^{j}+2r_k^{ij}),\\
\end{aligned}    
\end{equation}
where $\delta x_k = x_k - \bar{x}_k$, $\delta u_k^{j} = u_k^{j} - \bar{u}_k^{j}$, $Q_k^{i}, R_k^{ij}$ are the Hessians with respect to $x_k$ and $u_k^{j}$, and $l_k^{i}, r_k^{ij}$ are the gradients with respect to $x_k$ and $u_k^{j}$. We take the value of $\frac{I_{i,m}}{2}$ as the same as the nominal trajectory and treat it as a constant as an approximation when constructing the quadratic cost approximation. Same as \cite{ILQGames}, we omit the mix partial derivatives. 
\subsubsection{Separation Principle for Linear-Quadratic Stochastic Game}
With $(7)$ and $(22)$, we have constructed a linear-quadratic stochastic game. Next, we present the separation principle for linear-quadratic stochastic games, which would be exploited to solve the linear-quadratic stochastic game.

\begin{theorem}[Separation principle for linear-quadratic stochastic games]
With assumption 1, consider stochastic games with linear dynamics and quadratic costs for each player $i$:
\begin{equation*}
\begin{aligned}
x_{k+1}  &= A_kx_k+\sum_{j=1}^{N} B_k^ju^{j}_{k}+W_{k}w_{k},\\
y_{k+1}  &=  H_{k+1}x_{k+1} + V_{k+1}v_{k+1},\\
J_{i} &= \mathop{\mathbb{E}}_{X_0,...,X_L,Y_0,...,Y_L}\Bigg[\frac{1}{2}x_L^{T}(Q_L^{i}x_L+2l_L^{i})\\
&\quad+\sum_{k=0}^{L-1} \left(\frac{1}{2}x_k^{T}(Q_k^{i}x_k+2l_k^{i})+\frac{1}{2}\sum_{j=1}^{N}{u_k^{j}}^{T}(R_k^{ij}u_k^{j}+2r_k^{ij})\right)\Bigg],
\end{aligned}  
\end{equation*} 
where $w_k$ and $v_{k+1}$ are zero-mean Gaussian white noises. The state $x_0$ has a Gaussian prior $(\hat{x}_0^{p},\Sigma_{x_0}^{p})$, and the initial observation is $y_0 = H_0 x_0 + V_0 v_0$.

Assume $V_k \Sigma_{v_k} V_k^T \succ 0$ for all $k$, and $R_k^{ii} \succ 0$ for all $k$ and for each player $i$. Assume the game is well-posed, such that there exists a feedback Nash equilibrium for the associated deterministic linear-quadratic game and the second-order sufficient conditions for each player are satisfied.

Then a Nash equilibrium strategy of the above stochastic linear-quadratic game is a feedback Nash equilibrium strategy of the deterministic linear-quadratic game applied to the estimated state $\hat{x}_k$ provided by the Kalman filter.
\end{theorem}

The proof of Theorem 1 is in the appendix. Thanks to Theorem 1, we can obtain the Nash equilibrium strategy of the stochastic linear-quadratic game by solving the deterministic linear-quadratic game, which has an analytical solution \cite{bacsar1998dynamic}.


\subsection{Algorithm for Chance-Constrained Iterative Linear-Quadratic Stochastic Game}

The proposed algorithm is summarized in algorithm 1. After obtaining the Nash equilibrium, a line searching procedure is necessary to find a suitable step size. We use the definition of step size as in equation (7) in \cite{ILQGames}. We define the merit function to be $M =\frac{1}{2}\|\frac{\partial l^{i} }{\partial u^{i}}\|^{2}$ and apply a backtracking line search procedure to find a step size that satisfies the Armijo condition \cite{Nocedal2006Numerical}. We follow \cite{BeliefSpaceILQG,StochasticDynamicGames} to use a zero-noise realization to forward the dynamics. Then we use the extended Kalman filter \cite{ProbabilisticRobotics} to propagate the belief to obtain the next iteration's nominal trajectory.
\begin{algorithm}[htbp]

        \KwIn{Initial belief $b_0 = (\hat{x}_0,\Sigma_{x_0})$, nominal control $\bar{u}$}
        \KwOut{Belief trajectory $\bar{b}=\{\bar{b}_0,\bar{b}_1,...,\bar{b}_L\},\bar{u}$ and feedback control policy $\{\gamma_i\}, i \in \{1,...,N\}$}
        
        $\bar{b} \leftarrow $ Propagate belief with $\bar{u}$\;
      
        \While{max chance constraint error $>$ tolerance} 
        {
          Linearize chance constraints around $\bar{b}$ and evluate chance constraint violations\;
          update $\lambda,\mu$\;
          
        \While{not converge} 
        {
            Linearize the system and chance constraints around the nominal trajectory $\bar{b}, \bar{u}$\;
            Transform Lagrange multiplier and penalty terms, then add to the running costs\;
            Obtain a quadratic cost approximation\;
            $\{\gamma_i\}\leftarrow $Solve the linear-quadratic stochastic game\;
            Backtracking line search for step size $\alpha$\;
            $\bar{b}, \bar{u} \leftarrow $ Propagate belief with control policy $\{\gamma_i\} $\;

        }

        }
\caption{Chance-Constrained Iterative Linear-Quadratic Stochastic Game}
\end{algorithm}

\section{EXPERIMENTS}
\begin{figure*}[htbp]
    \centering
    \subfloat[]{\includegraphics[width=0.25\textwidth]{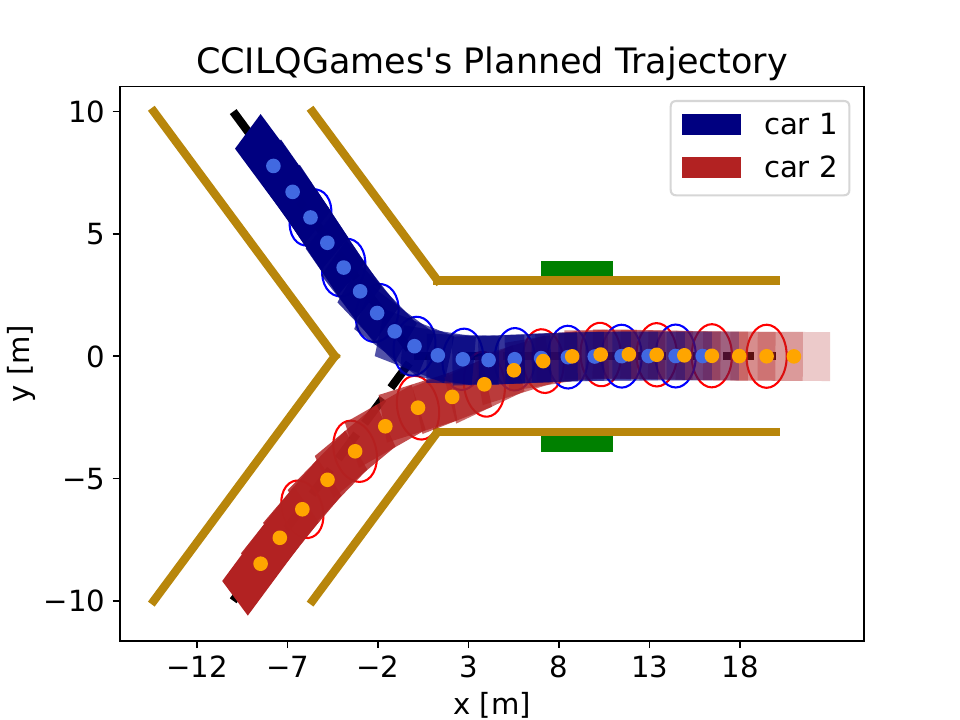}} 
    \subfloat[]{\includegraphics[width=0.25\textwidth]{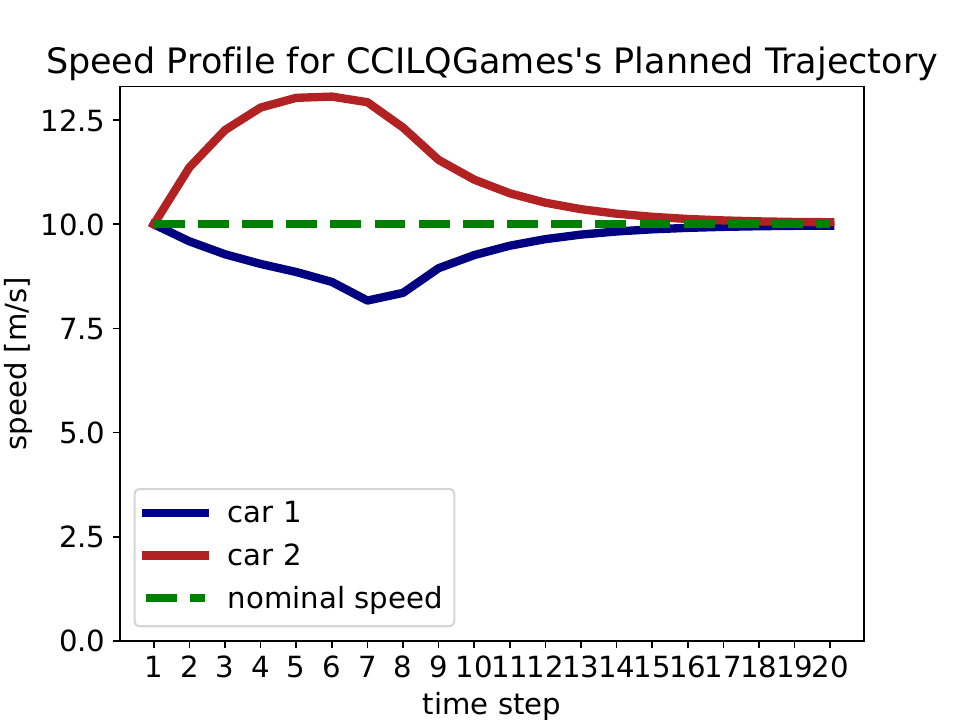}} 
    \subfloat[]{\includegraphics[width=0.25\textwidth]{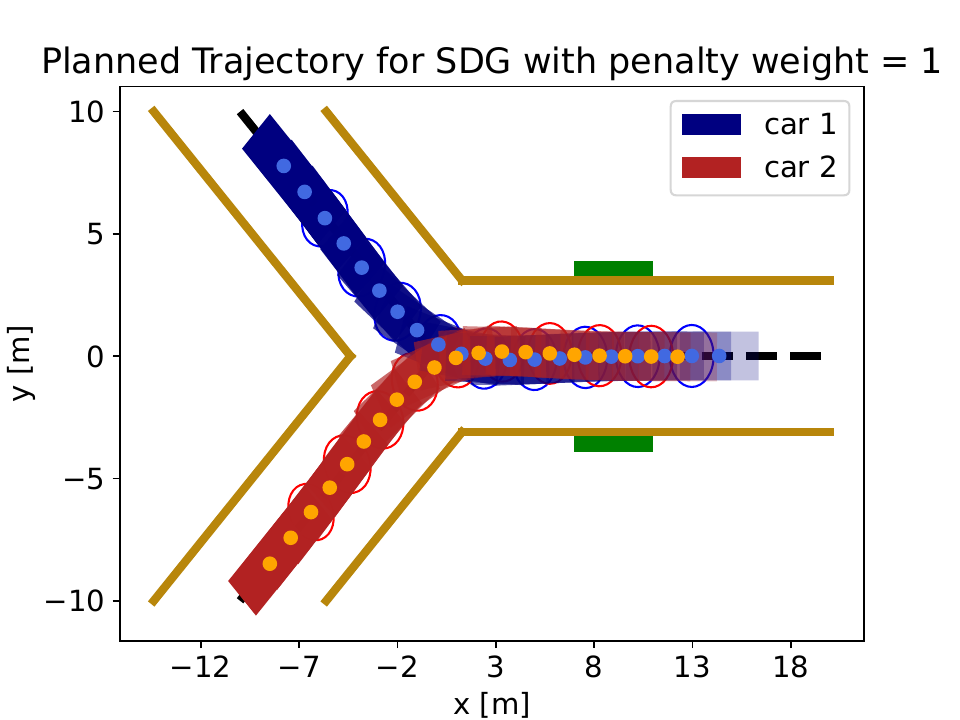}}
    \subfloat[]{\includegraphics[width=0.25\textwidth]{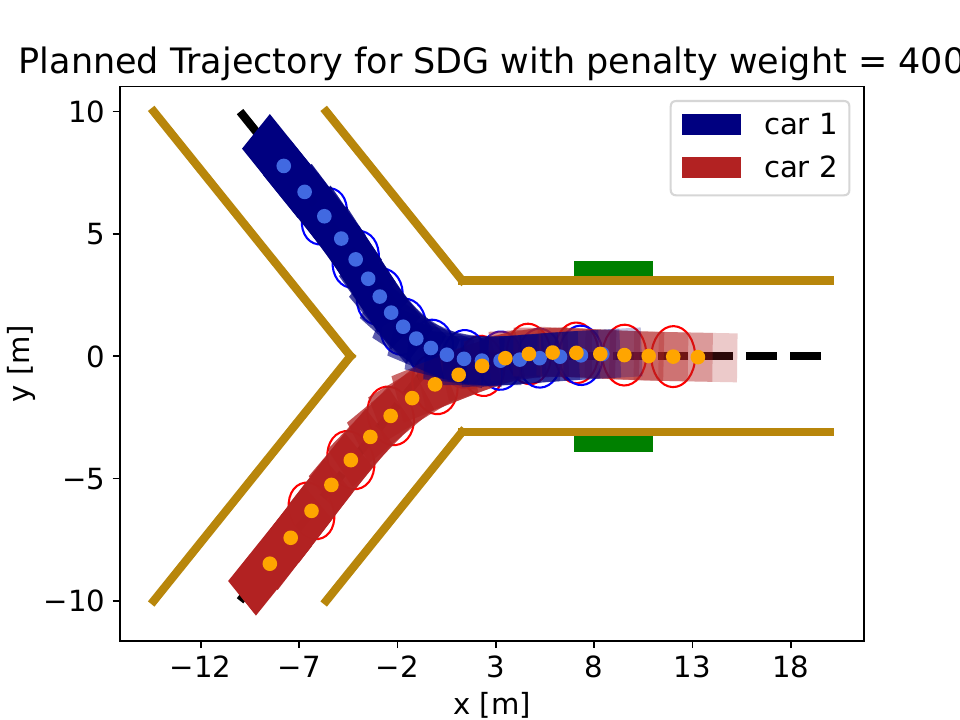}}
    \quad
    \subfloat[]{\includegraphics[width=0.25\textwidth]{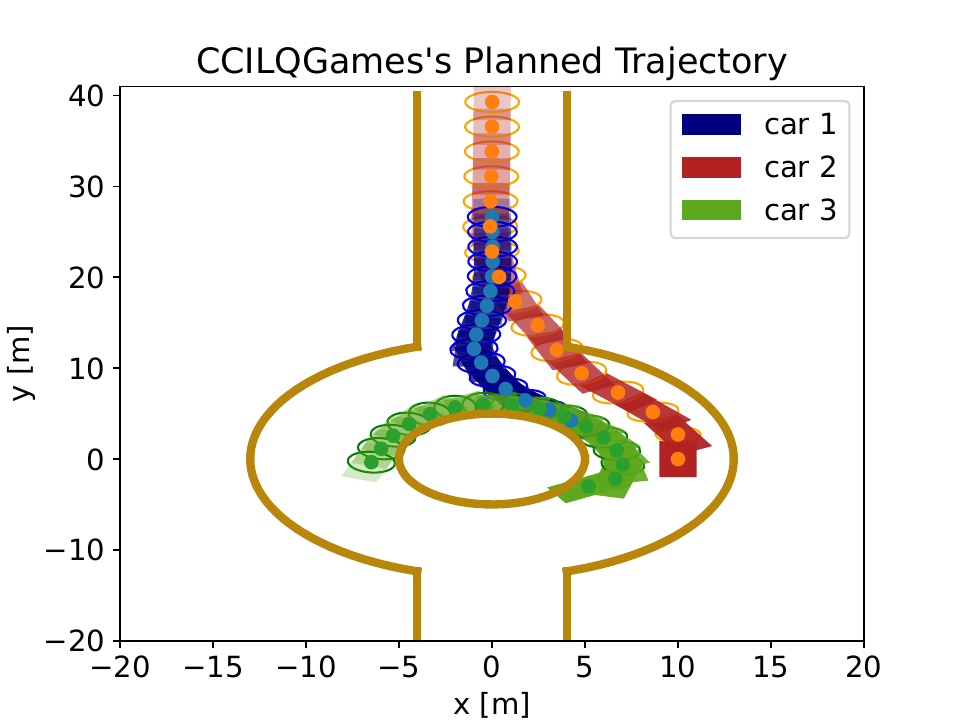}} 
    \subfloat[]{\includegraphics[width=0.25\textwidth]{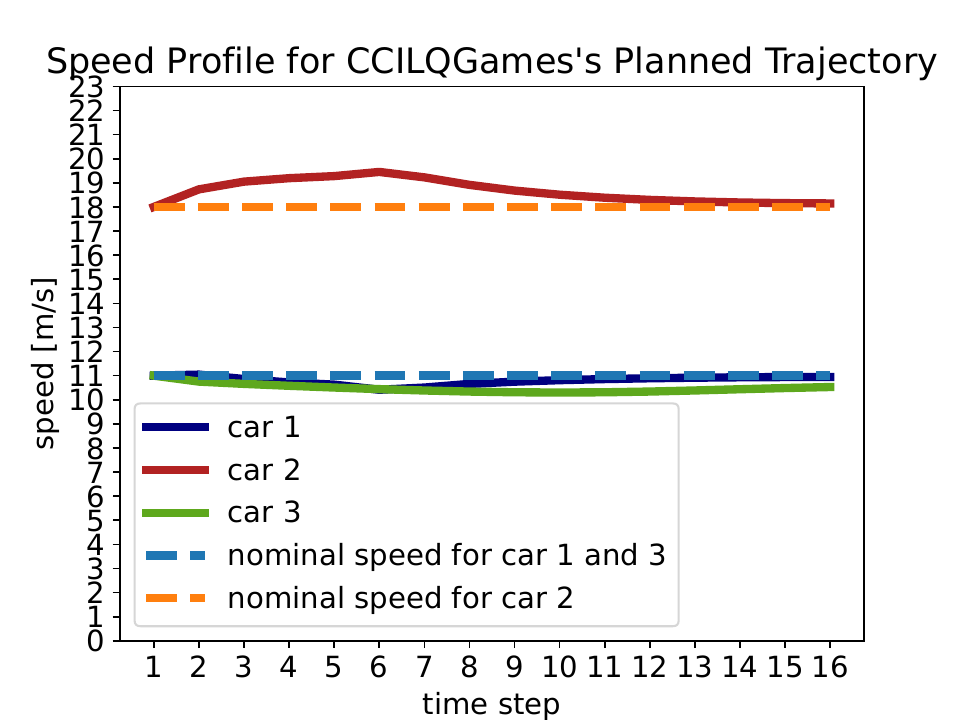}} 
    \subfloat[]{\includegraphics[width=0.25\textwidth]{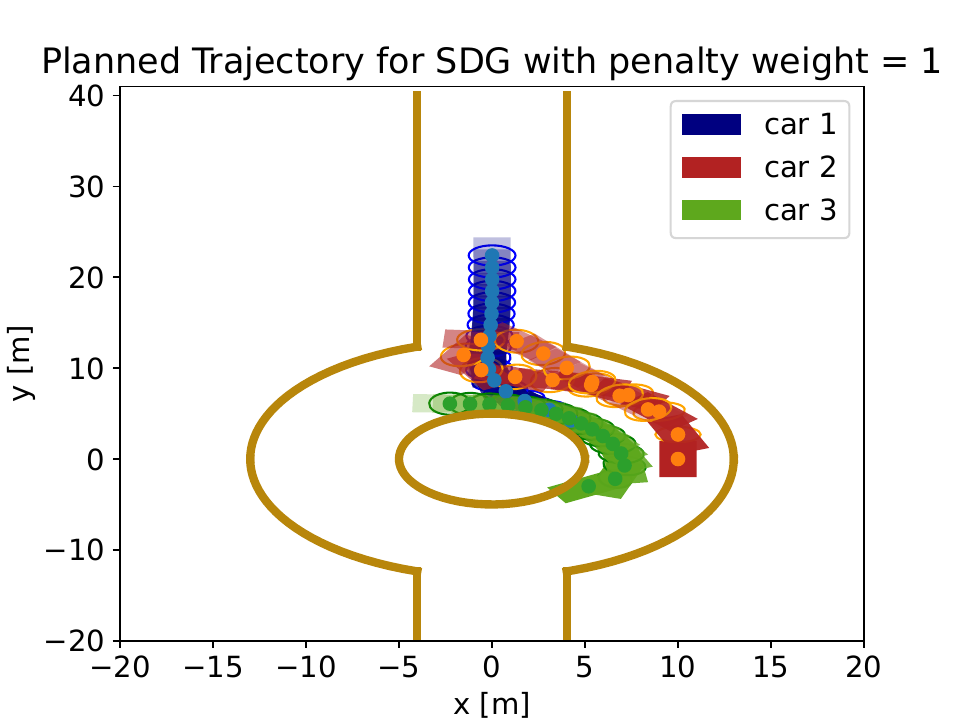}}
    \subfloat[]{\includegraphics[width=0.25\textwidth]{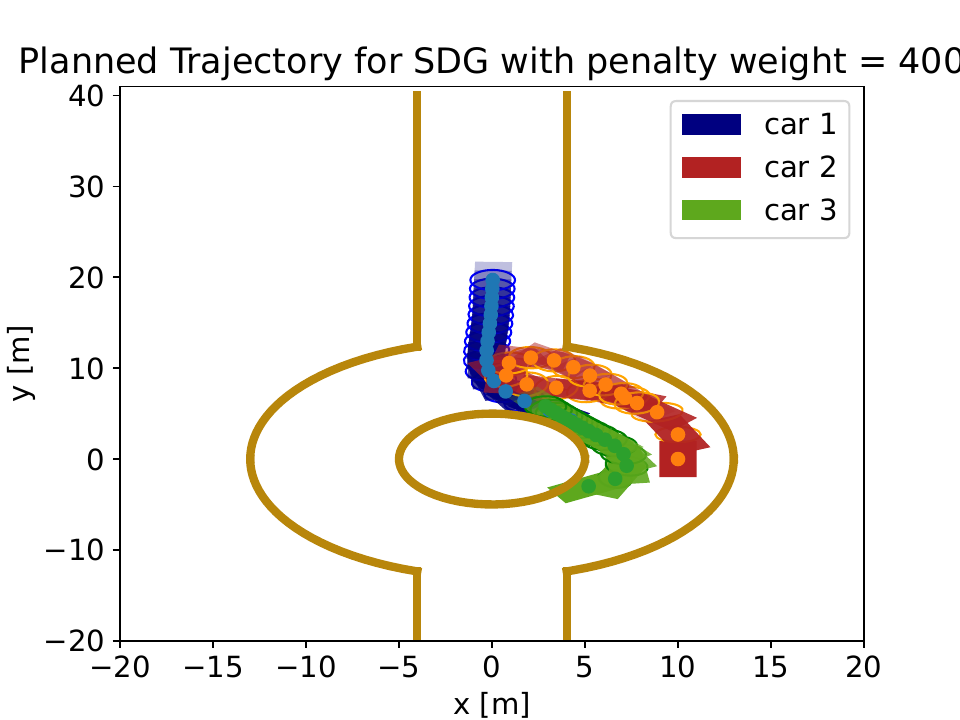}}
    \caption{The first row illustrates the results for the lane merging scenario, while the second row shows the results for the roundabout scenario.}
    \label{fig:LaneMerging Scenario and Roundabout Scenario}
\end{figure*}

\subsection{Experimental Setup}
In this section, we evaluate the performance of CCILQGames in three autonomous driving scenarios: lane merging, three-player intersection, and roundabout. We also further conduct a Monte Carlo study to test the proposed algorithm in noisy environments. We compare our algorithm to SDG with different constraint penalty weights. Since SDG is not open-sourced, we implement SDG by ourselves. All experiments are conducted on a 2.2 GHz Intel Core i7 laptop. We run CCILQGames 50 times for each scenario to calculate the computation time's mean and standard deviation. Computation time for lane merging, roundabout, and three-player intersection are $0.1017 \pm 0.007.3$ s, $0.4230 \pm 0.0212$ s, and $0.0866 \pm 0.0086$ s respectively. 

\subsubsection{Dynamics and Measurement Model}
We use the unicycle model as the vehicle dynamic model throughout the three scenarios. The state of the vehicle contains its position, heading angle, and velocity. The agent's control input is composed of its angular velocity and scalar acceleration.

We use the following measurement model for each agent in all three scenarios:
\begin{equation}
\begin{aligned}
{y}_{k+1}^{i} &= x_{k+1}^{i} + v_{k+1}^{i},\quad  v_{k+1}^{i} \sim \mathcal{N}(0,\Sigma_{v_{k+1}^{i}}), \\
\end{aligned}    
\end{equation}
where $x_{k+1}^{i}, {y}_{k+1}^{i}$ are the state and measurement of agent $i$ at time step $k+1$. $v_{k+1}^{i}$ is the associate noise. The joint measurement model is just the concatenation of all agents' measurement models.
\subsubsection{Costs and Chance Constraints}
We consider quadratic costs for distance from the lane center, deviation from the nominal speed, and control efforts, which are in the same form as in \cite{ILQGames}. We consider chance constraints on the proximity and collisions with obstacles. For the proximity constraints, the minimum distance threshold between two agents is set to 3 m. The obstacle avoidance constraint is formulated using the convex feasible set \cite{ConvexFeasibleSet,ChanceConstrainedILQG}. We augment the cost of the SDG simply with quadratic penalties for proximity constraint violations and linear penalties for obstacle avoidance constraint violation, all weighted by a penalty weight (1 or 400). We set the chance constraint threshold $p$ to 0.95 in all experiments.

\subsection{Lane Merging}
As shown in figure \ref{fig:LaneMerging Scenario and Roundabout Scenario},  two vehicles need to merge into the same lane in the lane merging scenario. Also, vehicles need to avoid collisions between each other and obstacles on the lane boundary. The planning horizon is 3 seconds with 20 time steps.

Figure \ref{fig:LaneMerging Scenario and Roundabout Scenario}(a) illustrates the planned trajectory using CCILQGames, in which the ellipses represent the covariance of the state. As demonstrated in the speed profile (Figure \ref{fig:LaneMerging Scenario and Roundabout Scenario}(b)), the red vehicle speeds up to cut into the lane ahead of the blue vehicle. By contrast, the blue vehicle slows down and allows the blue vehicle to merge. As shown in figure \ref{fig:LaneMerging Scenario and Roundabout Scenario}(c), a small penalty weight of 1 leads to unsafe strategies: two vehicles do not avoid collisions with each other. The Monte Carlo tests further validate a high probability of constraint violations. A large penalty weight of 400 leads the blue vehicle to slow down to let the red vehicle passes first. Notice that the penalty weight of 400 does not lead to a conservative strategy in this case. But the corresponding strategy is less safe than CCILQGames's strategy, as we would illustrate in the Monte Carlo Tests.
\begin{figure*}[htbp]
    \centering
    \subfloat[]{\includegraphics[width=0.1111\textwidth]{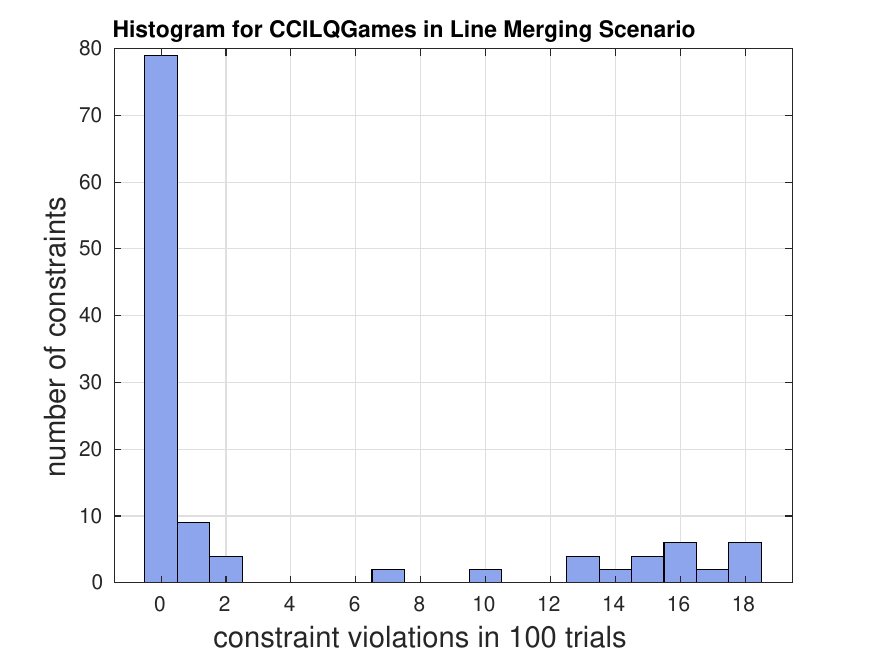}} 
    \subfloat[]{\includegraphics[width=0.1111\textwidth]{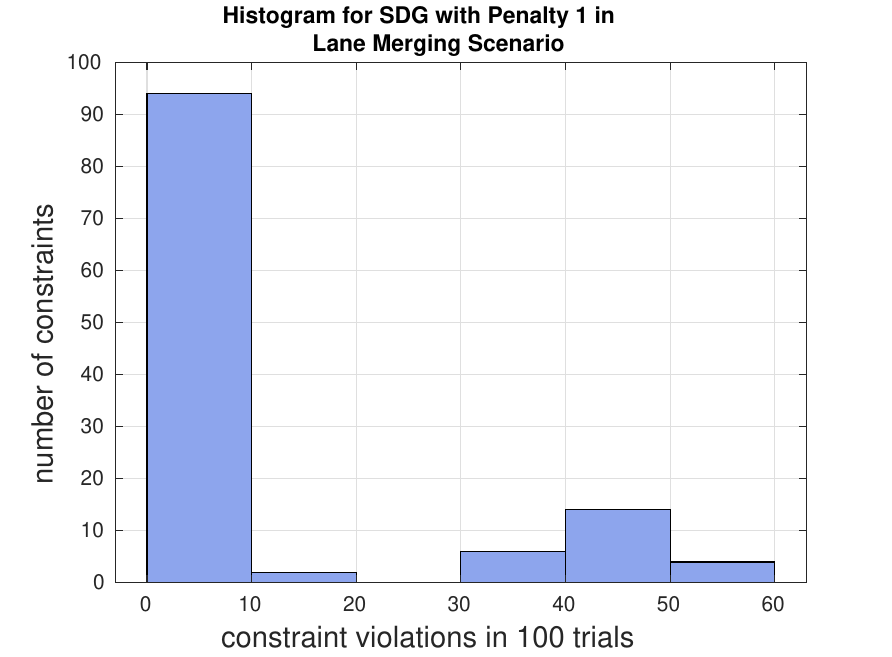}} 
    \subfloat[]{\includegraphics[width=0.1111\textwidth]{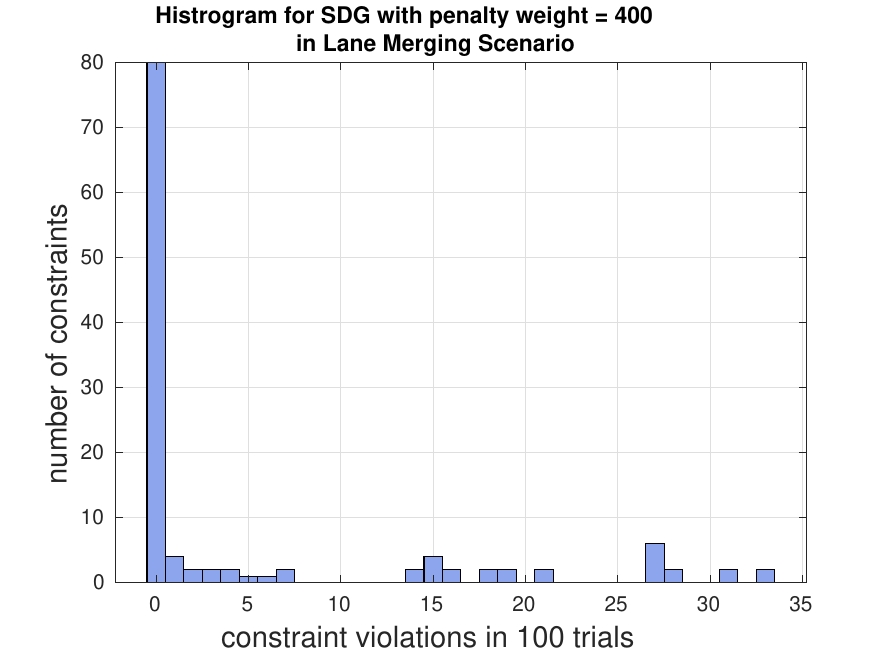}}
    \subfloat[]{\includegraphics[width=0.1111\textwidth]{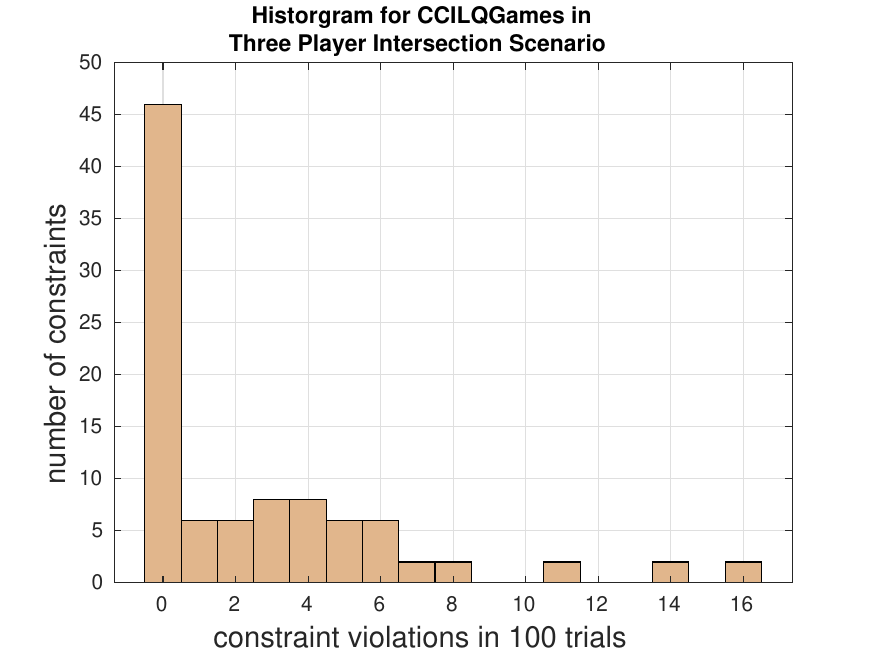}} 
    \subfloat[]{\includegraphics[width=0.1111\textwidth]{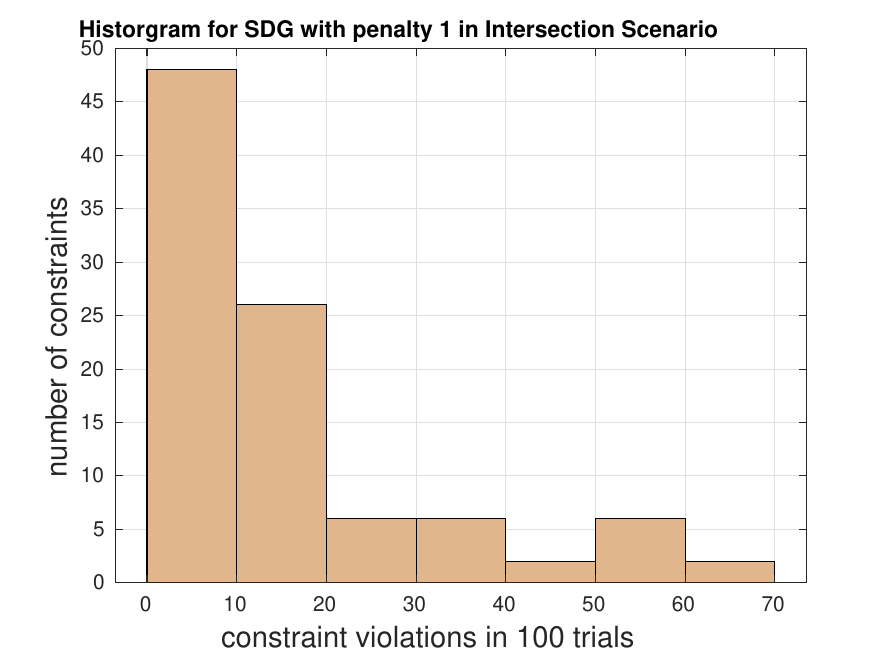}} 
    \subfloat[]{\includegraphics[width=0.1111\textwidth]{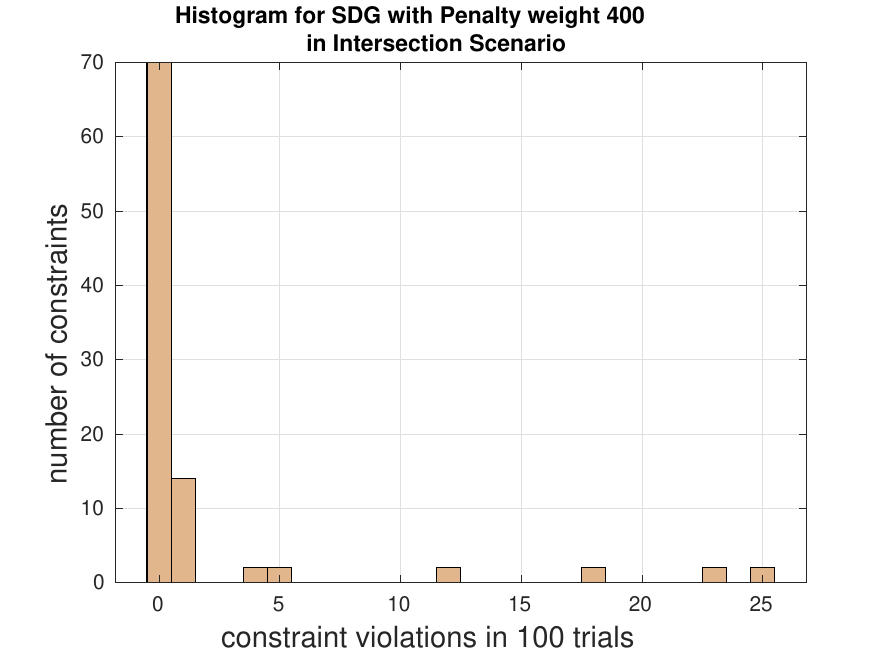}}
    \subfloat[]{\includegraphics[width=0.1111\textwidth]{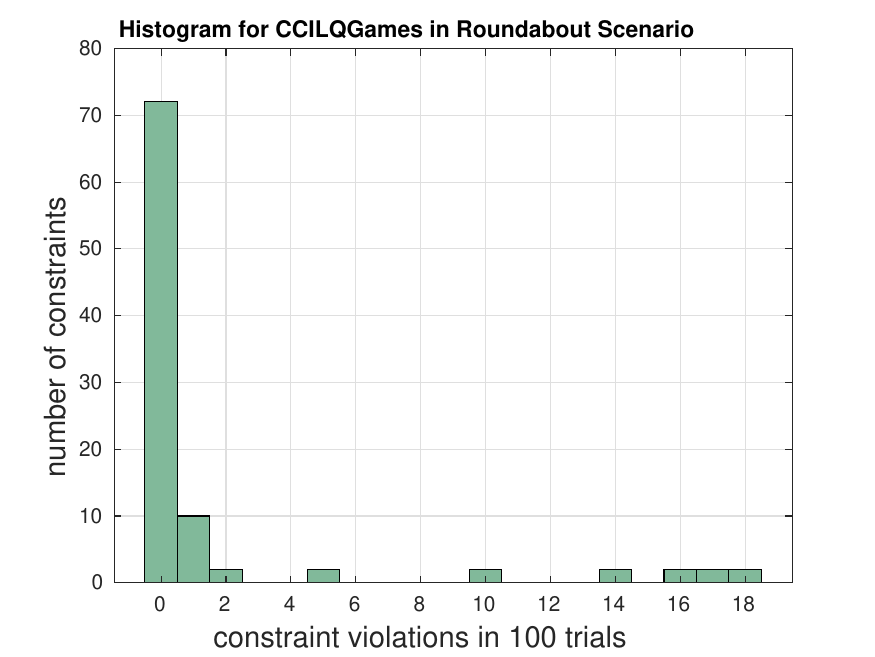}} 
    \subfloat[]{\includegraphics[width=0.1111\textwidth]{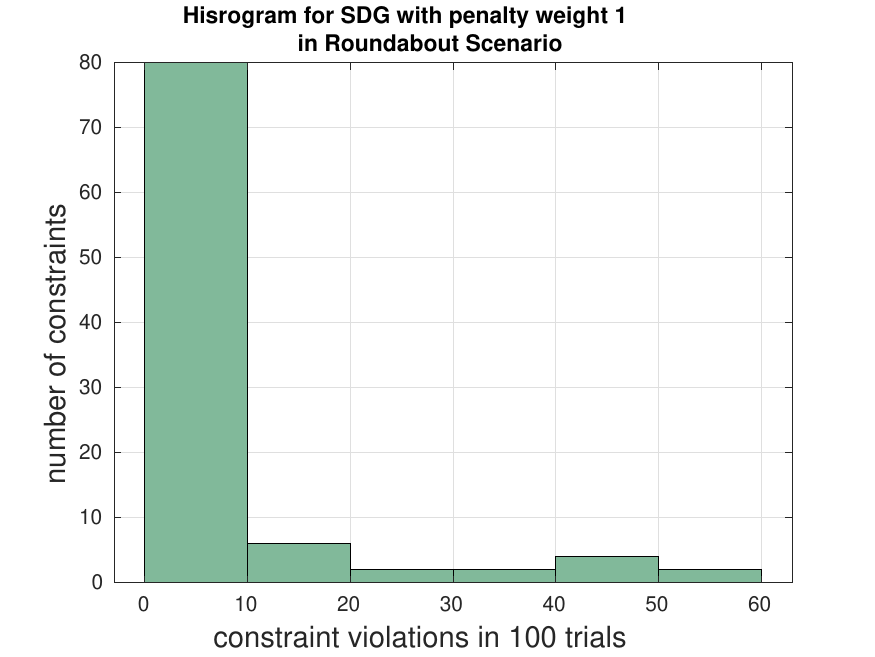}} 
    \subfloat[]{\includegraphics[width=0.1111\textwidth]{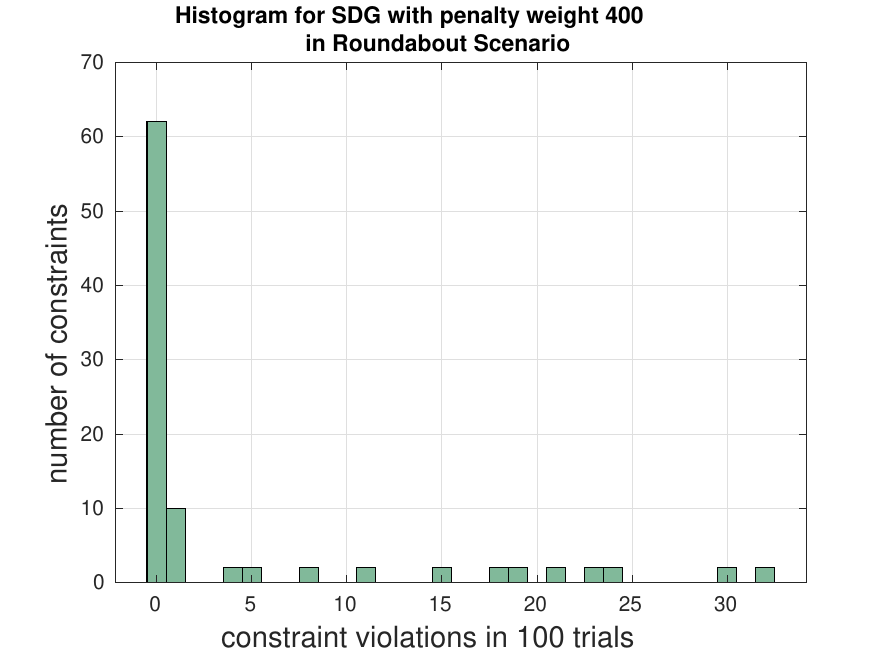}}
    \caption{Histograms of constraint violations for all the constraints in 100 Monte Carlo Tests, with sampled process and observation noises.}
    \label{fig:Histogram and Sample Trajectory}
\end{figure*}

\subsection{Three-Player Intersection}
In the three-player intersection scenario, the planning horizon is 2.5 s with  16 time steps. Figure \ref{fig:Motivating example}(a) illustrates the trajectory planned by CCILQGames. The red vehicle accelerates to pass the intersection ahead of the green vehicle. On the other hand, the blue vehicle slows down to wait for the red vehicle to go across the intersection. Interestingly, the red vehicle tweaks to avoid collisions. SDG with a small penalty weight could lead to unsafe behaviors. All vehicles go straight across the intersection without avoiding collisions (Figure \ref{fig:Motivating example}(c)). However, a large penalty weight again leads to conservative strategies, as only the red vehicle could pass the intersection while the others turn back (Figure \ref{fig:Motivating example}(d)).

\subsection{Roundabout}
The planning horizon is 2.5 s with  16 time steps for the roundabout scenario. Figure \ref{fig:LaneMerging Scenario and Roundabout Scenario}(e) illustrates the trajectory planned by CCILQGames. The red vehicle accelerates to escape the roundabout and cuts into the lane earlier than the blue vehicle. The blue vehicle's strategy is to slow down to let the red vehicle pass first. The green vehicle also slows down to leave space for the blue vehicle. Figure \ref{fig:LaneMerging Scenario and Roundabout Scenario}(g) shows the trajectory for SDG with a penalty weight of 1. The red vehicle turns around after escaping the roundabout since the cost only encourages to stay close to the lane center instead of specifying a direction. The strategy leads to collisions, which are validated by the Monte Carlo Tests. By comparison, a large penalty weight again leads to conservative behaviors, as only the blue vehicle navigates successfully (Figure \ref{fig:LaneMerging Scenario and Roundabout Scenario}(h)). The green vehicle slows down and could not get out of the Roundabout while the red vehicle turns back.

\subsection{Monte Carlo Tests}
We conduct a Monte Carlo study to test the performance of CCILQGames in stochastic environments. We run 100 trials for each scenario. The process noises are sampled from $\mathcal{N}(0,0.1I)$, with the exception of heading angle's variance being $0.05$. The observation noises are sampled from  $\mathcal{N}(0,0.6I)$, with the exception of angular velocity's variance being $0.1$. The noises' covariances are the same for planning (the planned trajecctoreis are shown in previous subsections) and Monte Carlo testing. Recall that our constraints are imposed on all the time steps in all three experiments. As shown in our problem formulation, we treat the same type of constraints at different time step as different constraints (e.g., the proximity constraint at time steps 1 and 2 are counted as two different constraints). Therefore, we calculate the statistics of constraint violations for each constraint in 100 trials. Figure 3 shows the histograms of constraint violations in 100 trials for CCILQGames and SDG in the three scenarios.

The maximum number of constraint violations among all the constraints for CCILQGames are 18,16,18 in lane merging, intersection and roundabout scenarios in 100 trials. By comparison, the maximum number of constraint violations among all the constraints for SDG with penalty weights 1(400) are 54 (33), 62 (25), 54 (22) in lane merging, intersection and roundabout scenarios in 100 trials. 

\subsection{Limitations}
We would like to discuss several limitations of CCILQGames. First, since we handle the chance constraints via linear approximations, there are no strict chance constraint satisfaction guarantees. Additionally, the linearization techniques could only specify a chance constraint threshold for each individual constraint instead of a joint threshold for satisfying all the constraints. As a result, CCILQGames achieves rates of 48\%, 51\%, and 51\% of constraint satisfaction for all the constraints in lane merging, intersection and roundabout scenarios in 100 trials. By comparison, SDG with the penalty weight 1 (400) achieves rates of 13\% (35\%), 6\% (65\%), 19\% (32\%) in lane merging, intersection and roundabout scenarios in 100 trials. We argue that the above limitations could be mitigated or resolved via an Model Predictive Control (MPC) formulation. For example, \cite{HaiZhuChance} uses the same technique to handle chance constraints in an MPC framework and achieves zero chance constraint violations in experiments. Another possible way to mitigate this limitation is to use the Bonferroni correction \cite{BonferroniCorrection} to assign a chance constraint threshold to each individual constraint. One more limitation of CCILQGames is that the Kalman filter could experience numerical instabilities in some cases (e.g., the covariance matrix becomes not symmetric positive definite).
\section{CONCLUSIONS}
We have presented a novel algorithm for solving stochastic dynamic games under chance constraints. Our work extends the deterministic ILQGames to stochastic games with both observation and process noises while handling chance constraints using the framework of augmented Lagrangian. We showcased the proposed algorithm in the lane merging, three-player intersection, and roundabout scenarios. The experimental results proved the effectiveness of the proposed approach.
\section*{APPENDIX}
\subsection{Proof for Theorem 1}
\begin{proof}
We can reformulate the cost function as follows:
\begin{equation}
\begin{aligned}
J_{i} &= \mathop{\mathbb{E}}_{X_0,...,X_L,Y_0,...,Y_L}[\frac{1}{2}x_L^{T}(Q_L^{i}x_L+2l_L^{i})\\
&+\sum_{k=0}^{L-1} (\frac{1}{2}x_k^{T}(Q_k^{i}x_k+2l_k^{i})+\frac{1}{2}\sum_{j=1}^{N}{u_k^{j}}^{T}(R_k^{ij}u_k+2r_k^{ij})) ],\\
&= \mathop{\mathbb{E}}_{Y_0,...,Y_L}[\frac{1}{2}\hat{x}_L^{T}(Q_L^{i}\hat{x}_L+2l_L^{i})\\ 
&+\sum_{k=0}^{L-1} (\frac{1}{2} \hat{x}_k^{T}(Q_k^{i}\hat{x}_k+2l_k^{i})+\frac{1}{2}\sum_{j=1}^{N}{u_k^{j}}^{T}(R_k^{ij}u_k+2r_k^{ij})) ]\\
&+\sum_{k=0}^{L} \frac{1}{2} trace(Q_k^{i}\Sigma_{x_k}),\\
\end{aligned}
\end{equation}
where $\hat{x}_k = \mathbb{E}[x_k \mid Y_{0:k}]$ is the conditional mean of the state at time step $k$. Notice that the trace terms $trace(Q_k^{i}\Sigma_{x_k})$ are constant and do not depend on control strategies. Due to assumption 1, the dynamics of $\hat{x}_k$ are the same for different agents. In other words, given the same observation-noise realizations, all agents obtain the same $\hat{x}_k$ from their Kalman filters (by Assumption 1, both the state measurement and the underlying observation-noise realization are shared across agents at each stage). Since $\hat{x}_k$ is fully observable (given by Kalman filter \cite{OcLinearQuadraticMethods}), the problem is reduced to a stochastic linear-quadratic game with closed-loop perfect state information pattern. By Corollary 6.4 in \cite{bacsar1998dynamic}, the Nash equilibrium of the linear-quadratic stochastic game with exact state information coincides with the deterministic version of the linear-quadratic game. Linear costs of states and controls are added into the cost function, which is different from the original setting of \cite{bacsar1998dynamic}. Although our regularity conditions differ from those in \cite{bacsar1998dynamic}, this conclusion also follows from a similar stage-wise dynamic-programming derivation on the transformed estimated-state system. Also, we do not claim the uniqueness of Nash equilibria in this setting.
\end{proof}







\bibliographystyle{IEEEtran}
\bibliography{hai_iros.bib}

\end{document}